\documentclass{article}
\usepackage{spconf,amsmath,epsfig}
\usepackage{algpseudocode}
\usepackage[ruled,vlined]{algorithm2e}
\usepackage[dvipsnames]{xcolor}
\usepackage{comment}
\usepackage{cleveref}
\usepackage{diagbox}
\usepackage{amsmath}
\usepackage{bm}
\usepackage{multirow}
\usepackage{slashbox}
\usepackage{makecell}

\usepackage{caption}
\usepackage{fancyhdr}

\fancypagestyle{copyright}{%
    \fancyhf{}
    \cfoot{\footnotesize%
        \textcopyright~2022 IEEE. Personal use of this
        material is permitted. Permission from IEEE must be obtained for all other
        uses, in any current or future media, including reprinting\slash republishing
        this material for advertising or promotional purposes, creating new collective
        works, for resale or redistribution to servers or lists, or reuse of any
        copyrighted component of this work in other works. 
    }
}

\let\OLDthebibliography\thebibliography
\renewcommand\thebibliography[1]{
  \OLDthebibliography{#1}
  \setlength{\parskip}{0pt}
  \setlength{\itemsep}{0pt plus 0.3ex}
}

\definecolor{commentcolor}{RGB}{110,154,155}   
\newcommand{\PyComment}[1]{\textcolor{commentcolor}{\# #1}}  
\newcommand{\PyCode}[1]{\textcolor{black}{#1}} 

\title{Self-supervised vision transformers for joint SAR-optical representation learning}
\name{Yi Wang$^{1,2}$, Conrad M Albrecht$^{1,2}$, Xiao Xiang Zhu$^{1,2}$\thanks{This work is supported by Helmholtz Association’s Initiative and Networking Fund through Helmholtz AI. Codes and pre-trained weights are available at https://github.com/zhu-xlab/DINO-MM.}}
\address{
$^1$Remote Sensing Technology Institute, German Aerospace Center (DLR), Germany \\
$^2$Data Science in Earth Observation, Technical University of Munich (TUM), Germany
}
\begin{document}
%
\maketitle

\begin{abstract}
Self-supervised learning (SSL) has attracted much interest in remote sensing and Earth observation
due to its ability to learn task-agnostic representations without human annotation. While most of the
existing SSL works in remote sensing utilize ConvNet backbones and focus on a single modality, we
explore the potential of vision transformers (ViTs) for joint SAR-optical representation learning.
Based on DINO, a state-of-the-art SSL algorithm that distills knowledge from two augmented views of
an input image, we combine SAR and optical imagery by concatenating all channels to a unified input.
Subsequently, we randomly mask out channels of one modality as a data augmentation strategy. While training,
the model gets fed optical-only, SAR-only, and SAR-optical image pairs learning both inner- and
intra-modality representations. Experimental results employing the BigEarthNet-MM dataset demonstrate
the benefits of both, the ViT backbones and the proposed multimodal SSL algorithm DINO-MM.   
\end{abstract}
\begin{keywords}
Self-supervised learning, vision transformer, multimodal representation learning
\end{keywords}

\thispagestyle{copyright}

\section{Introduction}
\label{sec:intro}

Recent advances in self-supervised learning (SSL) demonstrate great success in computer vision and remote sensing.
Designed to generate task-agnostic representations from large-scale, unlabeled data, SSL is attracting significant
attention due to an increasing amount of openly available earth observation data. A common SSL pipeline disassembles
into two major steps:
(1) based on a \textit{self-supervision} objective, a model is trained to learn high-level representations from massive amounts of unlabeled input data;
(2) the \textit{pre-trained} model from (1) is then transferred to a supervised \textit{downstream} task exploiting its ability to capture good representations.

The history of SSL yields various designs for self-supervision: (1) traditionally, generative methods like
autoencoders \cite{ballard1987modular} learn representations by reconstructing input data;
(2) in the past few years, predictive methods work by solving \textit{pretext tasks} such as predicting the relative position of two cropped patches \cite{doersch2015unsupervised};
(3) recently, contrastive methods utilize Siamese architectures to contrast the similarity between two augmented views of
the same input. A major benefit of contrastive learning is the freedom for the model to not depend on a specific pretext task,
but rather learn general representations. Yet, this can also lead to a trivial solution---the identity mapping.
Various methods have been designed to avoid this \textit{model collapse}, namely: negative sampling \cite{oord2018representation},
clustering \cite{caron2020unsupervised}, knowledge distillation \cite{chen2021exploring}, or redundancy reduction \cite{zbontar2021barlow}.
While most of recent SSL works in remote sensing are based on negative sampling \cite{jean2019tile2vec,ayush2021geography},
our approach picks DINO (self \textbf{di}stillation with \textbf{no} labels) \cite{caron2021emerging}, a state-of-the-art contrastive SSL method of category knowledge distillation.

Apart from the design of self-supervision, the choice of a backbone network is a key component to capture expressive representations.
ConvNets, ResNets \cite{he2016deep} in particular, are popular backbones in computer vision and remote sensing.
The recent success of vision transformers (ViTs) \cite{dosovitskiy2020image} in computer vision transfers to ViTs
as SSL model backbones \cite{caron2021emerging,chen2021empirical,he2021masked}.
Yet, the potential of self-supervised ViTs remains unlocked in remote sensing data analytics.
Our work targets to bridge this gap.

\begin{figure*}[ht]
    \centering
    \includegraphics[width=0.8\textwidth]{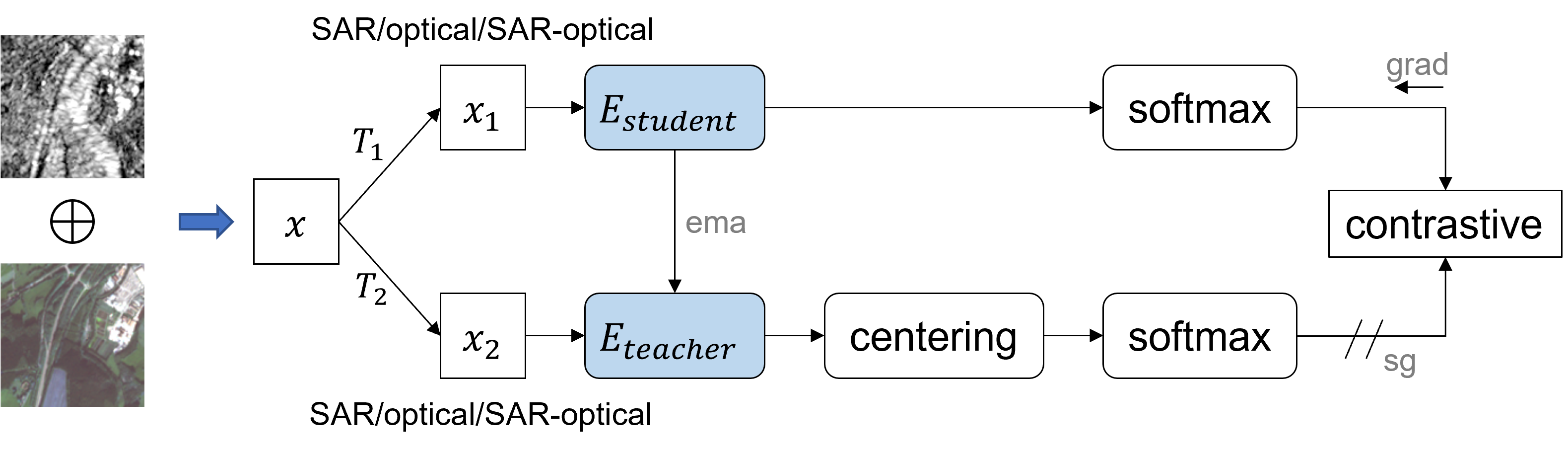}
    \caption{\textit{DINO-MM}: the proposed joint SAR-optical SSL algorithm. The concatenated SAR-optical image is taken
    as raw input. It is randomly transformed into two augmented views and fed into a DINO-based teacher-student
    network. The transformations $T_{1/2}$ include both regular data augmentations and a
    \textit{RandomSensorDrop} module that randomly masks out SAR or optical channels at a given probability.
    This way, the model contrasts views of either a single modality or both, learning both inner- and inter-modality
    representations. The abbreviation $ema$ denotes \textit{exponential moving average}, and $sg$ represents
    \textit{stop gradient}.}
    \label{fig:dino-mm}
\end{figure*}

Moreover, while a typical SSL algorithm considers a single modality, real-world problems usually require approaches
to jointly analyze multiple modalities. This is particularly true in earth observation where a variety of sensors
capture the physical properties of Earth's surface. Two popular modalities in remote sensing are synthetic-aperture
radar (SAR) and multispectral (optical) images. Although supervised SAR-optical fusion has been extensively studied,
corresponding self-supervised techniques are currently in their early stage \cite{cha2021contrastive,montanaro2021self,chen2021self0}. 
Meanwhile, it is also key to adjust those multimodal algorithms to situations where only a single modality is available.
Concerning these two aspects, we propose a joint SAR-optical representation learning method, DINO-MM, simultaneously able
to distill knowledge from a single or both modalities. With a newly proposed RandomSensorDrop module, the model randomly
receives SAR, optical, or concatenated SAR-optical images as input for contrastive learning. Therefore, the pre-trained
model is flexible to infer representations from any combination of available modalities.

\section{Methodology}
\label{sec:methodology}

\subsection{Vision transformers}


The vision transformer (ViT) \cite{dosovitskiy2020image} is a neural network model for image analysis.
It employs a Transformer-like architecture stacking attention blocks in order to draw global
dependencies of input sequences \cite{vaswani2017attention} over patches of an image.
Unlike in natural language processing (NLP) where a sentence is a sequence, an image needs
splitting into fixed-size patches. These patches are then linearly embedded to build the input sequence
for a ViT. After adding position embeddings, the resulting sequence of vectors is fed to a standard
Transformer encoder. For classification, a learnable \textit{classification token} gets added
to the sequence. Similar to ConvNets, ViTs may work as model backbones for SSL with the output
of the last layer's classification token identified as encoded representation vectors.

\subsection{DINO-MM}

Our work is based on DINO \cite{caron2021emerging}---a recent contrastive SSL algorithm with proven
performance when applied in conjunction with ViT backbones. We extend DINO to multimodal SSL, and
reference it by DINO-MM. It adds a simple data augmentation module, RandomSensorDrop, that randomly
masks out SAR or optical channels of the concatenated input image for joint multimodal representation learning.

\subsubsection{DINO}

Self-\textbf{di}stillation with \textbf{no} labels (DINO) is a knowledge distillation-based 
contrastive SSL algorithm maximizing the similarity of representations spawning from augmented views of the
same input image. \Cref{fig:dino-mm} depicts the model structure. An initial step generates two distorted views\footnote{%
    DINO follows a multi-crop strategy that invokes more than two views \cite{caron2020unsupervised}.
    To illustrate, we limit the discussion to two views.
} of an image, $x_{1}$ and $x_{2}$. These two views are sent to a Siamese teacher-student network,
$E_{student}$ and $E_{teacher}$. Both networks share the same architecture $g$ with separate sets of parameters
$\theta_s$ and $\theta_t$. The neural network $g$ builds on a backbone $f$ (e.g.\ a ViT encoder) and a
multi-layer perceptron (MLP) projection head $h$: $g = h \circ f$. The output of the teacher network gets
centered by a mean computed over the batch, which can be interpreted as adding a bias term $c$:
$g_{t}(x) \leftarrow g_{t}(x) + c$. The center $c$ is updated with an exponential moving average:
\begin{equation}
\label{eq:emaCenter}
c \leftarrow m c+(1-m) \frac{1}{B} \sum_{i=1}^{B} g_{\theta_{t}}\left(x_{i}\right),
\end{equation}
where $m$ is a rate parameter, and $B$ denotes the batch size. Each network outputs a $K$--dimensional
feature vector normalized by a temperature softmax, $\sim\exp(z_k/\tau)$, per feature dimension $z_k$
with $k=1\dots K$. In a final step, the similarity of the features from both networks is quantified
by a cross-entropy loss. 

DINO avoids model collapse by \textit{centering} the teacher network outputs through $c$, and by
\textit{sharpening} with a low teacher softmax temperature, i.e.\ it holds $\tau_t<\tau_s$ for the 
softmax normalizations. A stop-gradient operator applies to the teacher on training to make the
gradients backpropagate through the student network, only. Teacher parameters $\theta_t$ receive 
corresponding updates through an exponential moving average (\textit{ema}) of the student parameters
$\theta_s$---in analogy to \cref{eq:emaCenter}. Once self-supervised pre-training converged,
the network backbone $f$ serves as a representation generator to any downstream task.

\subsubsection{RandomSensorDrop}

We propose RandomSensorDrop as an additional data augmentation module to upgrade DINO for joint
SAR-optical representation learning. Consider a batch of SAR-optical image pairs with dimensions
$[B,C_{SAR/optical},W,H]$ where $B$ represents batch size, and $C_{SAR/optical}$ represents the
number of SAR or optical channels, respectively. For DINO-MM, we concatenate the pair of images
into a raw input $x$ with dimension $[B,C_{optical}+C_{SAR},W,H]$.
Once established, in addition to regular data augmentations of DINO not touching channel information,
we add the RandomSensorDrop module. This module randomly picks one of the \textit{three} options
for the \textit{two} views $x_1$ and $x_2$: replace SAR, optical, or none of the channel values by zero.

The procedure outlined enables the model to digest any possible combinations of the two modalities
as summarized by \Cref{tab:combinations}. Consequently, the model is expected to learn both
inner- and inter-modality representations:
\begin{itemize}
    \setlength\itemsep{.05em}
    \item $x_{1}$ and $x_{2}$ contain SAR images, only, i.e.\ the model learns SAR-only representations;
    \item $x_{1}$ and $x_{2}$ contain optical images, only, i.e.\ the model distills optical-only representations;
    \item $x_{1}$ and $x_{2}$ contain both modalities, i.e.\ the model learns joint SAR-optical representations.
\end{itemize}
\noindent As a result, the pre-trained model is able to infer data representations with either a single modality
or with both modalities available.

\begin{table}
\caption{Possible combinations of the two augmented views. 
}
\label{tab:combinations}
\centering
\scalebox{0.85}{

\begin{tabular}{|c|ccc|}
\hline
\backslashbox{~~\quad$\bm{x_1}$~~}{~~$\bm{x_2}$\quad~~}      & SAR (S) & Optical (O) & SAR-optical (M) \\ \hline
SAR (S)         & SS      & SO          & SM              \\
Optical (O)     & OS      & OO          & OM              \\
SAR-optical (M) & MS      & MO          & MM              \\ \hline
\end{tabular}
}
\vspace{-2ex}
\end{table}

\section{Experiments}
\label{sec:experiment}

\subsection{Dataset and implementation details}

\quad
\textbf{Dataset.} We evaluate DINO-MM on BigEarthNet-MM \cite{sumbul2021bigearthnet}---a
large-scale, multi-label scene classification dataset that contains 590,326 pairs of Sentinel-1 and Sentinel-2
image patches. We split the dataset into 311,667 training pairs, 103,944 validation pairs,
and 118,065 testing pairs. Patches got dropped where the scene is fully covered in seasonal
snow, or those including artifacts of clouds. All patches are re-sampled to 10m spatial
resolution such that each image has the size of 120x120 pixels.
We perform self-supervised pre-training on the training split without labels. 

\textbf{Model.} We use ViT-S/8 as backbone where $S$ indicates a small patch embedding dimension
of 384, and $/8$ signals input images get split into patches of size 8x8. We design the projection
head to a 3-layer MLP with hidden dimension 2048. It is followed by $l_2$ normalization and a
weight-normalized fully connected layer of dimension 65,536.

\textbf{Data augmentation.} We add the proposed RandomSensorDrop augmentation to the end of the existing
DINO augmentations including: RandomResizedCrop, RandomHorizontalFlip,
RandomColorJitter, RandomGrayscale, RandomGaussianBlur, and
RandomSolarize. We follow the multi-crop strategy with 2 global and 8 local crops.

\textbf{Self-supervised pre-training.} We train the model for 100 epochs applying the AdamW optimizer with batch size 256. The training takes about 8 hours on 4 NVIDIA A100 GPUs. The learning rate is linearly ramped up to $5\cdot10^{-4}$ over the first 10 epochs and then decayed with a cosine schedule. The temperature $\tau_s$ is set to $0.1$ and a linear warm-up increases $\tau_{t}$ from $0.04$ to $0.07$ over the first 30 epochs.

\textbf{Performance evaluation.}
We quantify performance of the proposed method by \textit{linear classification}, i.e., we train a
linear classifier on top of the parameter--frozen encoder $f$. We define the teacher backbone as
the pre-trained model. We train the linear classifier for 100 epochs with a batch size of 256,
SGD optimizer, and a cosine-decayed learning rate starting from $0.01$. As reference, we compare
the resulting performance to fully supervised learning running on the AdamW optimizer and a
cosine-decayed learning rate starting at $10^{-3}$.
The \texttt{torch.nn.Multi\-Label\-Soft\-Margin\-Loss} is applied for supervised training.
We report the performance score in terms of mean average precision.

\subsection{Results}

\begin{table}[t]
\centering
\caption{%
Linear classification results on the BigEarthNet-MM dataset
\cite{sumbul2021bigearthnet}. We report random initialization, self-supervised pre-training with a single
modality (DINO-S1/2), joint SAR-optical pre-training (DINO-MM), and fully supervised learning.
We also report the performance under label-limited regimes where only 1\% of the labels are available
to train the linear classifier.}
\label{tab:results}
\scalebox{0.9}{
\begin{tabular}{|c|ccc|ccc|}
\hline
\multirow{2}{*}{} & \multicolumn{3}{c|}{100\%}                    & \multicolumn{3}{c|}{1\%}                      \\ \cline{2-7} 
                  & S1            & S2            & S1+S2         & S1            & S2            & S1+S2         \\ \hline
Random            & 54.6          & 62.0          & 64.5          & 52.7          & 59.0          & 62.4          \\
DINO-S1/2         & 76.2          & 86.0          & --            & 68.7          & 82.0          & --            \\
DINO-MM           & \textbf{79.5} & \textbf{87.1} & 87.1          & \textbf{75.3} & \textbf{82.9} & \textbf{82.8} \\
Supervised        & 77.1          & 86.7          & \textbf{88.6} & 63.7          & 73.6          & 75.0          \\ \hline
\end{tabular}}
\vspace{-2.5ex}
\end{table}
\Cref{tab:results} verifies the success of self-supervised ViTs and highlights the benefits of DINO-MM.
While table rows represent different training models, corresponding columns list
downstream tasks with full labels (100\%) and a fraction of labels (1\%), respectively.

\textbf{Self-supervised ViTs.} In contrast to random initialization (row 1), self-supervised pre-training
significantly boosts performance by a margin of $\ge20$\% points (rows 2, 3). This result underlines the
capacity of ViTs to generate comprehensive representations for remote sensing data---an observation verified
when compared to supervised learning (row 4): training a linear classifier with 100\%
labels, ViT-based SSL reaches performance close to fully supervised learning.

When reducing the fraction of labels to 1\%, self-supervised pre-training outperforms supervised learning
on any combination of Sentinel-1/2 imagery. This result advertises SSL as an efficient approach in situations
with a limited amount of labels available. In addition, while the performance score for supervised learning
significantly drops with number of labels reduced ($\ge13$\% points), the performance of linear classification
based on pre-trained models is affected little ($\le5$\% points). This demonstrates the benefits of SSL to
help minimize human annotation efforts.

\textbf{Joint multimodal SSL.} A comparison of SSL based on a single modality, and SSL with both modalities
is presented in rows 2 and 3 of \Cref{tab:results}, respectively. With both modalities (DINO-MM on S1+S2),
the performance is improved compared to a single modality (DINO-S1 on S1 or DINO-S2 on S2)---a result expected
since an additional modality adds information. Further, the results suggest optical imagery
capture more relevant information than SAR raster data for the specific scene classification task at hand.
Most significantly, \Cref{tab:results} supports the benefits of joint multimodal SSL: a model pre-trained with both
modalities transfers well to single-modality tasks, and it can even boost performance. That is, DINO-MM on S1 or S2
outperforms DINO-S1 on S1 or DINO-S2 on S2, and it even outperforms supervised learning with all the labels.
This result is critical to remote sensing applications such as early change detection for disaster monitoring
where either one or the other satellite captures a scene first.

\section{Conclusion}
\label{sec:conclusion}

In this paper, we explored the potential of self-supervised vision transformers (ViTs) for multimodal remote
sensing image understanding. Based on a state-of-the-art self-supervised learning (SSL) algorithm, DINO, we proposed
and evaluated a model extension, DINO-MM, to jointly learn representations of SAR-optical data. We introduced a 
simple data augmentation module, RandomSensorDrop, that randomly masks out channels of one modality while training.
Experimental results verify the success of ViT backbones for SSL in remote sensing and prove the benefits
of the proposed multimodal SSL method.


\small{
\bibliographystyle{IEEEtran}
\bibliography{refs}
}
\end{document}